\documentclass[10pt]{article}
\ifdefined\pdfminorversion\pdfminorversion=7\fi
\usepackage{icassp2027_paperkit}
\usepackage{amsmath,amssymb,booktabs,graphicx,microtype,url}
\usepackage[T1]{fontenc}
\usepackage{textcomp,balance,caption}
\usepackage[hidelinks]{hyperref}
\hypersetup{pdftitle={Beyond Mean Foils: Auditing Worst-Foil Specificity in Frozen CLIP Region Explanations},pdfauthor={Kaixin Liu; Zhipeng Ye; Feng Jiang; Zhenghao Wang; Qihang Wu},pdfkeywords={CLIP, explanation auditing, fixed-population foil controls, repair capacity}}

\newcommand{\estci}[2]{\begin{tabular}[c]{@{}r@{}}#1\\{#2}\end{tabular}}
\newcommand{\foils}{\mathcal F}
\newcommand{\regions}{\mathcal R}
\title{Beyond Mean Foils: Auditing Worst-Foil Specificity in Frozen CLIP Region Explanations}
\name{\href{https://orcid.org/0009-0005-5213-8081}{Kaixin Liu}$^{1,*}$, \href{https://orcid.org/0000-0002-3384-2779}{Zhipeng Ye}$^{1,*,\dagger}$, \href{https://orcid.org/0000-0001-5362-3234}{Feng Jiang}$^{1}$, \href{https://orcid.org/0009-0002-8768-4937}{Zhenghao Wang}$^{1}$, \href{https://orcid.org/0009-0009-6082-0223}{Qihang Wu}$^{1}$}
\address{$^{1}$Taizhou Institute of Science and Technology, Nanjing University of Science and Technology,\\
Taizhou 225300, Jiangsu, China\\
\texttt{24107880127@nustti.edu.cn, zhipengye@nustti.edu.cn, jf@nustti.edu.cn,}\\
\texttt{wangzhenghao2002@outlook.com, 24107880128@nustti.edu.cn}\\
$^{*}$Equal contribution. $^{\dagger}$Corresponding author: Zhipeng Ye.}

\begin{document}
\maketitle
\begin{abstract}
A region can overlap a target object yet contribute more to another class.
We test regions selected by Cluster-based Concept Importance (CCI) in frozen CLIP. Across COCO and VOC with two
checkpoints, 41.08--64.78\% of regions that pass overlap and mean-contrast
checks fail against the strongest competing class. Removing competitors
annotated in the image leaves 39.69--63.64\% failing. We then test all eight candidate regions per image. An alternative passes the test for 6.25--7.84\% of failures on
COCO and 27.40--31.15\% on VOC. Requiring it to preserve the original
target-score drop within $\epsilon=.02$ reduces these rates to 0.16--0.98\%.
Available regions and target-drop tolerance constrain repair; relaxing
the tolerance increases repair opportunities.

\end{abstract}
\AddToHook{env/equation/begin}{
\setlength{\abovedisplayskip}{6pt plus 1pt minus 1pt}
\setlength{\belowdisplayskip}{6pt plus 1pt minus 1pt}
\setlength{\abovedisplayshortskip}{3pt plus 1pt}
\setlength{\belowdisplayshortskip}{4pt plus 1pt minus 1pt}
}

\section{Introduction}
Cluster-based Concept Importance (CCI)~\cite{agarwal2025cci} explains
CLIP~\cite{radford2021clip} by grouping patches into regions and measuring
how masking them changes image--text scores. We examine CCI-top1, the
region whose removal most reduces the target score.

A region may contribute more to another class than to the target, even
when its target drop exceeds the average competing drop. We test the
strongest competitor, or worst foil, and ask whether another region can
pass this test while preserving the target drop.

Class-contrastive explanations~\cite{wang2022whynot} can choose a
competitor by its score in the original image. Our audit instead chooses
the class whose score drops most when a region is masked.
Related work covers corpus foils~\cite{lin2023cocoa}, class-sensitive
saliency~\cite{williamson2025case}, CLIP
attribution~\cite{wang2023m2ib,li2023clipsurgery,dreyer2025cliplatent},
class contrast~\cite{li2026cdaclip}, and caption foils~\cite{yuksekgonul2023aro}.
We control for competing objects and foil count, then examine all regions
to locate constraints on repair.

\begin{figure*}[t]
\centering
\includegraphics[width=\textwidth]{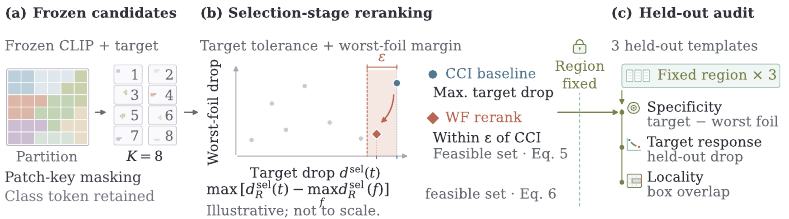}
\caption{Selection and held-out audit. A region is fixed using selection-stage
responses before three-template evaluation of specificity, target response,
and locality. Schematic responses illustrate the selection rule.}
\label{fig:method}
\end{figure*}

\section{Audit and Repair Opportunity}
Figure~\ref{fig:method} shows candidate construction (a), one-prompt
selection (b), and three-prompt evaluation of the fixed region (c).

For image $x$, let $t$ be its target and $c$ any class. The candidate set
$\regions(x)$ contains $K=8$ patch regions $R$. Checkpoint $m$ gives
pre-softmax score $s_m(x,c)$, including its learned logit scale.
Masking attention to $R$ gives $\mathcal I(x,R)$. The response $d_R(c)$
is the original score minus the masked score; positive values mean
a reduction:

\begin{equation}
d_R(c)=s_m(x,c)-s_m(\mathcal I(x,R),c).
\label{eq:drop}
\end{equation}
Superscript $\mathrm{sel}$ denotes ``a photo of a \{category\}''.
CCI-top1 chooses $R_{\rm CCI}$, the region with the largest target drop;
$\arg\max$ returns the maximizing region:
\begin{equation}
R_{\rm CCI}=\arg\max_{R\in\regions(x)}d_R^{\rm sel}(t).
\label{eq:cci}
\end{equation}

\paragraph{Evaluating class specificity.}
We evaluate the selected region with ``a picture of the \{category\}'',
``an image containing a \{category\}'', and ``the \{category\}''.
Index the $H=3$ prompts by $h$. Their unit-norm text embeddings are
$e_h(c)$ and their drops are $d_{R,h}(c)$. The mean embedding has length
$n_c=\|H^{-1}\sum_h e_h(c)\|_2$, where $\|\cdot\|_2$ is Euclidean length.
Dividing the mean drop by this length gives the drop for a normalized
mean embedding: $\bar d_c(R)=H^{-1}\sum_h d_{R,h}(c)/n_c$.

A foil is a non-target class, and $\foils_t$ is the set of all such classes.
With $f$ indexing foils and $R$ implicit in the drops, margin $P(R,t)$
subtracts the largest foil drop from the target drop. Negative values
mean that a foil loses more score:

\begin{equation}
P(R,t)=\bar d_t-\max_{f\in\foils_t}\bar d_f.
\label{eq:estimands}
\end{equation}
We call $P<0$ a failure and a transition to $P\ge0$ a sign repair.
Target-response tables use the raw mean drop $H^{-1}\sum_h d_{R,h}(t)$.

\noindent\begin{minipage}{\columnwidth}
For baseline CCI-top1, $A$ contains images passing both checks: bbox
precision (the fraction of patch centers inside target boxes) is at least
$.5$, and the target drop exceeds the mean foil drop. For image $i$ with
target $t_i$, write $P_i(R)=P(R,t_i)$; $P_i$ denotes its baseline margin.
Failures form $F=\{i:P_i<0\}$, so the screened failures are $B=A\cap F$.
These image sets stay fixed. The subset with positive target contribution
is $A^+=A\cap\{\bar d_t>0\}$.
\end{minipage}\par

\paragraph{Selecting an alternative region.}
Worst-foil (WF) reranking allows a target-drop loss of at most
$\epsilon\ge0$ relative to CCI-top1 (Figure~\ref{fig:method}(b)).
Its selection margin $M^{\rm sel}$ subtracts the largest foil drop
from the target drop under the selection prompt:
\begin{equation}
M^{\rm sel}(R,t)=d_R^{\rm sel}(t)-\max_{f\in\foils_t}d_R^{\rm sel}(f),
\end{equation}
The feasible set $\regions_\epsilon$ contains candidates within the
allowed target-drop loss:
\begin{equation}
\regions_\epsilon=\{R\in\regions(x):d_R^{\rm sel}(t)\ge d_{R_{\rm CCI}}^{\rm sel}(t)-\epsilon\},
\label{eq:feasible}
\end{equation}
WF chooses $R_{\rm WF}$ to maximize that margin within the feasible set:
\begin{equation}
R_{\rm WF}=\arg\max_{R\in\regions_\epsilon}M^{\rm sel}(R,t).
\label{eq:wf}
\end{equation}
We use $\epsilon=.02$ logit units and sweep $.01/.02/.05/.10/.20$.

\paragraph{Locating the constraints on repair.}
An oracle tests whether any allowed alternative can repair a failure.
It counts images in $B$ with a feasible region
with a nonnegative held-out margin. Denote this count by $N_O(B,\epsilon)$
and image $i$'s feasible set by $\regions_\epsilon(i)$. The indicator
$\mathbf1\{\cdot\}$ is one when its condition holds and zero otherwise:
\begin{equation}
N_O(B,\epsilon)=\sum_{i\in B}\mathbf1\!\left\{
\max_{R\in\regions_\epsilon(i)}P_i(R)\ge0\right\}.
\label{eq:oracle}
\end{equation}
The unrestricted oracle removes the budget constraint and searches all
$K$ candidates. Each image in $B$ falls into one of four groups: no
candidate has $P\ge0$ ($C_0$); a candidate passes but none meets the
budget ($C_1$); a feasible candidate passes but WF misses it ($C_2$);
or WF repairs the failure ($r$).
These groups partition $B$, whose size is $|B|$, so
$|B|=C_0+C_1+C_2+r$. The oracle counts both feasible groups: $N_O=C_2+r$.
The joint condition $J$ additionally requires overlap and mean-foil checks
and $\bar d_t>0$; $j$ counts WF repairs satisfying it within fixed $B$.
WF selects using its selection prompt; only the oracle uses held-out scores.

\paragraph{Controlling for competing objects.}
A foil may name another object in the image. To measure its contribution,
remove all annotated-present classes from image $i$'s foil set, leaving
$\foils_i^{\rm abs}$. The protocol holds $R_{\rm CCI}$ and full-foil screen $A$ fixed;
the failure event is $E_i^{\rm abs}=\{\bar d_t-\max_{f\in\foils_i^{\rm abs}}\bar d_f<0\}$.
Removing any foils can reduce failure, so we compare this exclusion with
randomly retaining the same number of foils. For $N_i$ full
foils, $k_i=|\foils_i^{\rm abs}|$, and $m_i$ foils with $\bar d_f>\bar d_t$,
let $q_i$ be the failure probability of a uniform size-$k_i$ subset.
With $\binom{n}{k}$ counting size-$k$ subsets, one minus the fraction
containing no stronger foil gives
\begin{equation}
q_i=1-\frac{\binom{N_i-m_i}{k_i}}{\binom{N_i}{k_i}}.
\label{eq:matched}
\end{equation}
The numerator is zero when $k_i>N_i-m_i$. To isolate the effect of
annotation-based removal, $\Delta_A$ averages its failure indicator minus
the random-subset probability over the $|A|$ baseline images:
$\Delta_A=|A|^{-1}\sum_{i\in A}(\mathbf1\{E_i^{\rm abs}\}-q_i)$.
Negative $\Delta_A$ favors annotation-based removal. Retained foil sets are nonempty.

\section{Protocol and Reproducibility}
We evaluate 27,708 COCO val2014~\cite{lin2014coco} and 1,943 VOC2007
test~\cite{everingham2010voc} images with CLIP B/16 and B/32 (16- and
32-pixel patches): 59,302 image--model records. The target is the largest
instance by annotation area (COCO) or box area (VOC). Images require another category and 2--80\% target area;
COCO crowd annotations are excluded.
Full foils are the other 79/19 classes.

K-means uses normalized projected patch features, $K=8$, three starts,
50 iterations, and seed $1701+i$; $i$ indexes sorted IDs shuffled with seed 1701. Intervention $\mathcal I$
adds $-\infty$ to attention logits at selected patch-key columns before
softmax in every block/head, retaining the class-token column. Ties use first argmax.
Boxes follow the 224-pixel short-side resize and center crop~\cite{gomez2022metrics}.

\clearpage
\section{Results and Analysis}
\noindent\begin{minipage}{\columnwidth}
\centering
\captionof{table}{Fixed baseline $A$: full-foil, annotation-absent, and exact cardinality-matched random failure rates (\%). $100\Delta_A$ is absent minus random in percentage points, with paired 95\% CI.}
\label{tab:screens}
\small\renewcommand{\arraystretch}{1.15}\setlength{\tabcolsep}{1.1pt}
\begin{tabular*}{\columnwidth}{@{\extracolsep{\fill}}lrrrr@{}}
\toprule
Setting & Full & Absent & Random & $100\Delta_A$\\
\midrule
COCO B/16 & 64.37 & 63.17 & 64.09 & \estci{-0.92}{[-1.08,-0.77]}\\
COCO B/32 & 64.78 & 63.64 & 64.50 & \estci{-0.86}{[-1.01,-0.71]}\\
VOC B/16 & 42.27 & 40.94 & 41.46 & \estci{-0.52}{[-1.16,+0.06]}\\
VOC B/32 & 41.08 & 39.69 & 40.24 & \estci{-0.55}{[-1.21,+0.04]}\\
 
\bottomrule
\end{tabular*}
\end{minipage}\par\vspace{6pt}

\paragraph{Removing annotated competitors leaves most failures.}
Table~\ref{tab:screens} compares the same baseline images before and after
removing classes annotated in the image. The set $A$ contains 17,726,
17,848, 1,209, and 1,227 images in table order. Failure falls by only
1.14--1.39 percentage points: 96.63--98.24\% of the failures in $B$ remain.
On COCO, exclusion helps more than removing the same number of classes
at random. On VOC, the confidence intervals for that difference include
zero. Annotated competing objects thus explain only a small part of the
observed failure.

The pattern also holds when the target contribution is positive. In
$A^+$, failure after exclusion remains 61.38--62.62\% on COCO and
39.26--40.05\% on VOC. Even COCO images with exactly two annotated
categories have failure rates of 63.39\% and 64.95\% for B/16 and B/32,
respectively. The mismatch therefore persists beyond crowded scenes.

\par\addvspace{6pt}
\noindent\begin{minipage}{\columnwidth}
\centering
\captionof{table}{Exact local decomposition of \emph{screened failures} $B=A\cap F$
at $\epsilon=.02$. Counts in $C_0,C_1,C_2,r$ are mutually exclusive;
$C_2+r$ is the feasible-oracle count.}
\label{tab:oracle}
\small\renewcommand{\arraystretch}{1.15}\setlength{\tabcolsep}{1.4pt}
\begin{tabular*}{\columnwidth}{@{\extracolsep{\fill}}lrrrrr@{}}
\toprule
Setting & $|B|$ & $C_0$ & $C_1$ & $C_2$ & $r$\\
\midrule
COCO B/16 & 11410 & 10697 & 695 & 0 & 18\\
COCO B/32 & 11562 & 10655 & 876 & 1 & 30\\
VOC B/16 & 511 & 371 & 135 & 0 & 5\\
VOC B/32 & 504 & 347 & 154 & 1 & 2\\
 
\bottomrule
\end{tabular*}
\end{minipage}\par\vspace{6pt}

\paragraph{Few failures have an alternative within the budget.}
Table~\ref{tab:oracle} shows where repair stops. For 92.16--93.75\% of
COCO failures and 68.85--72.60\% of VOC failures, none of the eight
candidates passes the worst-foil test ($C_0$). Figure~\ref{fig:ident} shows
the remaining opportunity as blue circles: 6.25--7.84\% on COCO and
27.40--31.15\% on VOC. Imposing $\epsilon=.02$ removes 96.43--98.09\%
of these opportunities, leaving only 0.16--0.98\% of $B$ repairable
(orange diamonds). WF misses at most one of these feasible repairs per
setting ($C_2$). The main losses occur before the final ranking decision.

\newpage
\par\addvspace{6pt}
\noindent\begin{minipage}{\columnwidth}
\centering
\includegraphics[width=\columnwidth]{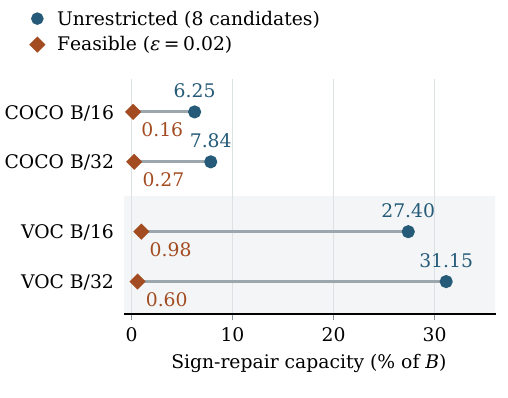}
\captionof{figure}{Budget-constrained sign-repair capacity among fixed screened failures $B$. Connected markers compare the same setting on a shared linear scale.
Unrestricted: at least one of all eight candidates has $P\ge0$; feasible:
at least one also meets $\epsilon=.02$. Percentages use each setting's own
$|B|$ (Table~\ref{tab:oracle}) and are computed by exact enumeration.}
\label{fig:ident}
\end{minipage}\par\vspace{6pt}

\par\addvspace{6pt}
\noindent\begin{minipage}{\columnwidth}
\centering
\captionof{table}{Budget endpoints (C: COCO; V: VOC). Multi: percentage of all eligible images with multiple feasible candidates. $O_B,r_B$: oracle/actual sign-repair percentages of screened failures $B$. $\Delta d_t,\Delta b$: all-image raw-target and bbox changes (bbox in percentage points).}
\label{tab:frontier}
\small\renewcommand{\arraystretch}{1.15}\setlength{\tabcolsep}{1pt}
\begin{tabular*}{\columnwidth}{@{\extracolsep{\fill}}lrrrrrr@{}}
\toprule
Set & $\epsilon$ & Multi & $O_B$ & $r_B$ & $\Delta d_t$ & $\Delta b$\\
\midrule
C16 & 0.02 & 4.71 & 0.16 & 0.16 & +0.0001 & -0.12\\
C16 & 0.20 & 35.78 & 1.57 & 1.50 & -0.0140 & -1.63\\
C32 & 0.02 & 4.40 & 0.27 & 0.26 & -0.0001 & -0.08\\
C32 & 0.20 & 34.95 & 2.05 & 2.00 & -0.0143 & -0.60\\
V16 & 0.02 & 4.79 & 0.98 & 0.98 & -0.0008 & -0.34\\
V16 & 0.20 & 34.48 & 6.65 & 6.07 & -0.0223 & +0.08\\
V32 & 0.02 & 4.73 & 0.60 & 0.40 & -0.0007 & +0.02\\
V32 & 0.20 & 34.64 & 7.14 & 6.75 & -0.0161 & +1.07\\
 
\bottomrule
\end{tabular*}
\end{minipage}\par\vspace{6pt}

\paragraph{A larger tolerance opens up more alternatives.}
At $\epsilon=.02$, about 95\% of images have only one feasible candidate
(Table~\ref{tab:frontier}). Increasing the tolerance to $.20$ gives multiple
choices on about 35\% of images. Oracle repair capacity rises to
1.57--2.05\% of COCO failures and 6.65--7.14\% of VOC failures.
The oracle-repair percentage $O_B$ and achieved WF percentage $r_B$
are computed as $O_B=100N_O/|B|$ and $r_B=100r/|B|$.

The extra choice changes what the selected region captures. The table's
$\Delta d_t$ and $\Delta b$ are WF minus CCI changes averaged over all
eligible images; bbox changes are in percentage points. At $.20$, raw
target drops decrease in all four settings, while overlap decreases on
COCO and increases on VOC. More repair opportunities therefore do not
imply a uniform improvement in target response or localization.

\clearpage
\par\addvspace{6pt}
\noindent\begin{minipage}{\columnwidth}
\centering
\captionof{table}{Normalization sensitivity on fixed earlier-run $A_{\rm old}$
and earlier-run CCI regions. Raw/Norm: failure rates (\%); Flip: percentage
whose failure status differs. $A_{\rm old}$ denotes earlier-run screen-passers.}
\label{tab:normalization}
\small\renewcommand{\arraystretch}{1.18}\setlength{\tabcolsep}{1pt}
\begin{tabular*}{\columnwidth}{@{\extracolsep{\fill}}lrrrr@{}}
\toprule
Setting & $|A_{\rm old}|$ & Raw & Norm & Flip\\
\midrule
COCO B/16 & 17,726 & 64.318 & 64.369 & 0.107\\
COCO B/32 & 17,847 & 64.762 & 64.778 & 0.073\\
VOC B/16 & 1,209 & 42.184 & 42.266 & 0.083\\
VOC B/32 & 1,227 & 41.157 & 41.157 & 0.000\\
 
\bottomrule
\end{tabular*}
\end{minipage}\par\vspace{3pt}

\paragraph{Normalization rarely changes the outcome.}
Table~\ref{tab:normalization} isolates normalization by fixing earlier-run
regions and images; these records are not pooled with rerun results.
Failure rates differ by at most 0.083 percentage points, and at most
0.107\% of images switch between passing and failing. The worst foil
is unchanged on 99.42--99.84\% of images. The high failure rate
therefore persists under both score definitions.

\par\addvspace{3pt}
\noindent\begin{minipage}{\columnwidth}
\centering
\captionof{table}{Actual sign repairs \emph{within fixed baseline $B$}, $\epsilon=.02$.
$r$ counts repairs; $j$ counts those also satisfying $J$.
WF$-$CCI means with repair-subset 95\% CIs. $\Delta d_t$ is raw
held-out target-score change; only bbox change is in percentage points.}
\label{tab:costs}
\small\renewcommand{\arraystretch}{1.10}\setlength{\tabcolsep}{1.1pt}
\begin{tabular*}{\columnwidth}{@{\extracolsep{\fill}}lrrrr@{}}
\toprule
Setting & $r$ & $j$ & $\Delta d_t$ & $100\Delta$bbox\\
\midrule
COCO B/16 & 18 & 12 & \estci{0.029}{[-0.096,0.148]} & \estci{-19.53}{[-38.12,-1.32]}\\
COCO B/32 & 30 & 25 & \estci{0.094}{[0.036,0.148]} & \estci{-14.04}{[-27.70,-1.74]}\\
VOC B/16 & 5 & 3 & \estci{-0.021}{[-0.170,0.182]} & \estci{-16.87}{[-26.11,-7.00]}\\
VOC B/32 & 2 & 1 & \estci{-0.078}{[-0.115,-0.042]} & \estci{-28.17}{[-78.57,22.22]}\\
 
\bottomrule
\end{tabular*}
\end{minipage}\par\vspace{3pt}

\paragraph{Repairing specificity can reduce spatial overlap.}
Table~\ref{tab:costs} considers only repaired failures; Table~\ref{tab:frontier}
uses all eligible images. Mean bbox precision falls by 14.04--28.17
percentage points among repairs. Its interval is below zero in three
settings, but spans zero for VOC B/32. Mean target-score changes are
positive on COCO and negative on VOC; their intervals span zero for both
B/16 settings. Thus the direction of a mean change is not always resolved
by its interval. VOC estimates summarize just five and two repairs.

A repaired margin also need not retain the other checks. On COCO B/32,
30 failures are repaired, but only 25 retain overlap, mean-foil contrast,
and positive target contribution. This gap between repairs ($r$) and
repairs passing all checks ($j$) occurs in every setting. Even the oracle
finds only 12, 26, 3, and 2 feasible repairs passing all checks, in table
order. Among all baseline passes on COCO B/32, four instead become
worst-foil failures; these regressions lie outside the fixed failure set.

\newpage
\par\addvspace{3pt}
\noindent\begin{minipage}{\columnwidth}
\centering
\captionof{table}{Nonzero-threshold sensitivity on fixed baseline $A$.
Entries are percentages of screen-passers $A$ with $P<-\delta$,
where $\delta\ge0$ is the required failure magnitude in the normalized margin.
Thresholds are evaluated post hoc in logit units.}
\label{tab:threshold}
\small\renewcommand{\arraystretch}{1.15}
\begin{tabular*}{\columnwidth}{@{\extracolsep{\fill}}lrrrr@{}}
\toprule
Setting & $\delta=0$ & $.01$ & $.10$ & $.50$\\
\midrule
COCO B/16 & 64.37 & 63.94 & 59.49 & 36.05\\
COCO B/32 & 64.78 & 64.38 & 60.22 & 35.19\\
VOC B/16 & 42.27 & 41.44 & 36.23 & 13.73\\
VOC B/32 & 41.08 & 40.26 & 34.39 & 11.90\\
 
\bottomrule
\end{tabular*}
\end{minipage}\par\vspace{6pt}

\paragraph{The failures are not confined to margins near zero.}
Table~\ref{tab:threshold} tightens the test from $P<0$ to $P<-\delta$.
At $\delta=.10$, 59.49--60.22\% of COCO screen-passers and
34.39--36.23\% of VOC screen-passers still fail. Rates decline further
at $.50$. Thus many failures survive a stricter margin requirement,
although their prevalence depends on the chosen threshold.

\paragraph{Implications for region selection.}
The comparisons point to two distinct changes. Table~\ref{tab:oracle}
identifies images that need a better candidate region ($C_0$) and images
that already have a passing region outside the budget ($C_1$).
Figure~\ref{fig:ident} shows how sharply the budget narrows the available
choices, and Table~\ref{tab:frontier} shows the effect of relaxing it.
Changing the ranking rule alone can recover only the additional feasible
opportunities in $C_2$. Improving repair therefore requires attention to
which regions are proposed and how much target-drop loss is allowed.

\paragraph{Reproducibility and uncertainty.}
Main results use a fully recorded rerun of the earlier seed rule, not
independent base-seed repetitions. Checks reconstruct regions, scores,
choices, and failures from saved masks, responses, and bbox records.
\href{https://github.com/jovial-liu/conway/tree/7ee2b6080aa6752790e17c910c5c8a3111235d4d/experiments/priority123_20260915}{Released records} include summaries, checks, configuration, and hashes;
masks and full responses remain local. Table~\ref{tab:normalization}
uses earlier-run records. Selected regions changed in three images.

We obtain confidence intervals from 10,000 image-level bootstrap samples,
using the 2.5th and 97.5th percentiles without multiplicity correction
or coverage of base-seed variability.
Paired effects are differenced within each image. Foil comparisons use
fixed $A$ or $A^+$; repair-cost intervals use only actual repairs.
We use exact $q_i$ values, checked against 100 seeded random draws.

\section{Conclusion}
A region can overlap the target and beat the average competing class while
contributing more to a single competitor. This occurs frequently for
CCI-top1, even after annotated competitors are removed. Checking every
candidate shows why reranking repairs so few failures: many images have
no passing alternative, and most remaining alternatives exceed the
target-drop tolerance. Relaxing that tolerance makes more repairs possible
but also changes target response and overlap. Worst-foil comparisons
therefore add information that overlap and mean contrast alone do not
provide when evaluating frozen CLIP region explanations.

\clearpage
\balance

\section*{Compliance with Ethical Standards}
This work analyses publicly released COCO and PASCAL VOC images and
annotations. No participants were recruited and no new human or animal
experiments were conducted.
\section*{Acknowledgments}
This work was supported by the Young Scientific and Technological Talent Support Program under the Taizhou Fengcheng Talent Plan.

The authors declare that they have no known competing financial interests or personal relationships that could have appeared to influence the work reported in this paper.
\end{document}